\documentclass[10pt,twocolumn,letterpaper]{article}

\usepackage[pagenumbers]{wacv} % To force page numbers, e.g. for an arXiv version

\usepackage{newunicodechar}
\newunicodechar{₁}{$_1$}
\usepackage{stfloats}
\usepackage{float}  % In your preamble if not already present
\usepackage{caption}
\usepackage{graphicx}
\usepackage{multicol}     % Already implicitly used in your style
\usepackage{lipsum}        % Optional for dummy text

\usepackage{amsmath}

\definecolor{wacvblue}{rgb}{0.21,0.49,0.74}
\usepackage[pagebackref,breaklinks,colorlinks,allcolors=wacvblue]{hyperref}

\def\wacvPaperID{*****} % *** Enter the WACV Paper ID here
\def\confName{WACV}
\def\confYear{2026}

\title{Improving Negation Understanding in Medical Vision–Language Models via Contrastive Fine-Tuning}

\author{Jasmine Vu\\ 
Santa Clara University\\
{\tt\small jqvu@scu.edu}
\and
Shivanand Sheshappanavar\\
University of Wyoming\\
{\tt\small ssheshap@uwyo.edu}
}

\begin{document}
\maketitle
\begin{abstract}
Large vision-language models like CLIP are increasingly used in medical imaging tasks due to their ability to align images and text without the need for extensive labeled data. This makes them particularly useful for applications like image retrieval, report generation, and classification in clinical settings. A potential issue to this approach is that CLIP-based models often under perform when interpreting negated phrases, which is especially problematic in the context of medical diagnosing. In this study, we evaluate the Stanford AIMI CheXagent model on its ability to correctly retrieve chest X-ray images using prompts with and without negation. The goal of this project is to understand where this model fails and then use it as a base model to improve its retrieval accuracy by fine tuning methods outlined in previous work. Results from this study show improvement in handling of negation in the CLIP model with a slight decrease in accuracy of positive prompt evaluation. Alongside retrieval accuracy, we examined internal model behavior through token attribution, t-SNE projection, and attention-head ablation to better characterize how each fine-tuning approach reshaped the text encoder’s representation of negated clinical language. Through this work, we hope to better understand the internal behavior of CLIP and improve its handling of negation using clinically relevant language for improving its reliability in medical AI devices. 
\end{abstract}
    
\section{Introduction}
\label{sec:intro}

Contrastive Language-Image Pretraining (CLIP) \cite{Authors14f} aligns text prompts with images through its inner text and image encoders, which generate tokens and pair them using cosine similarity. Through training across large datasets, its accurate alignment of texts and images can be improved and thus used for zero-shot performance. This allows people to input text prompts and retrieve relevant images, which is especially useful in healthcare, where clinicians may need to retrieve medically relevant images for reports and diagnoses. The ability to handle this retrieval without explicit labels makes CLIP and similar models appealing for clinical use \cite{Authors14g}, however, their use in medical settings can be difficult due to their struggle to recognize negation in text prompts (Figure 1). 

\begin{figure}
    \centering
    \includegraphics[width=\linewidth]{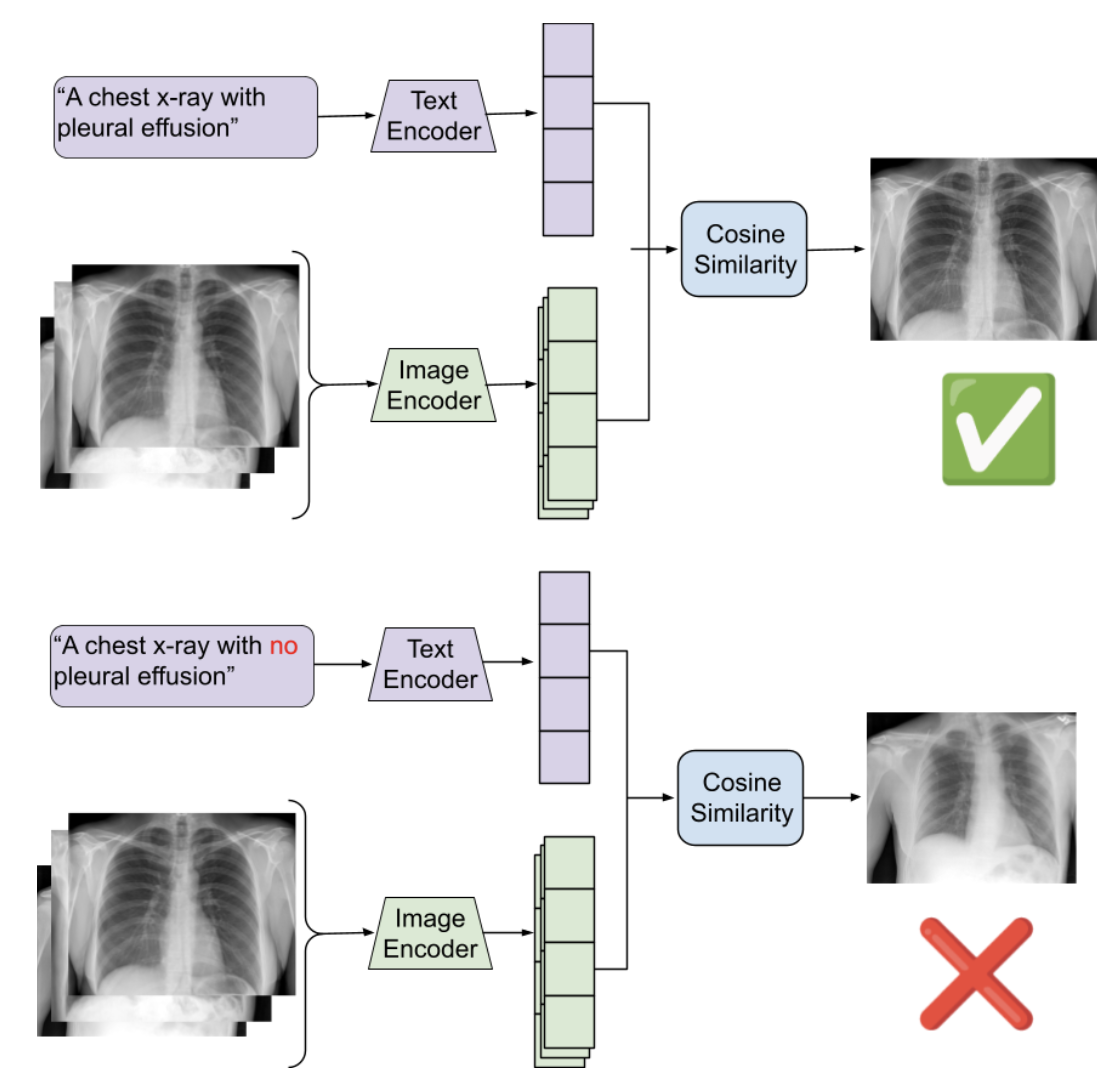}
    \caption{Flowchart of Stanford AIMI CheXagent chest X-ray image retrieval given a prompt containing affirmation for pleural effusion and a prompt containing negation for plueral effusion. All chest X-ray images shown in figures were obtained from the Open-i repository and originate from the PubMed Central Open Access subset or the Indiana University Chest X-ray dataset. All images are publicly available, fully de-identified, and used in accordance with their respective Creative Commons or open-access licenses\cite{Authors14e}.
}
    \label{fig:reunotes2}
\end{figure}

While there have been many adaptations of CLIP for use in medical settings, like CXR-CLIP \cite{Authors14i}, GLORIA \cite{Authors14j}, MedKlip \cite{Authors14l}, BioViL \cite{Authors14k}, MLIP \cite{Authors14m}, Xlip \cite{Authors14h}, they often overlook the importance of ensuring accuracy in the context of negation. Negative terms often appears in radiology reports to indicate the absence of a condition, and misinterpreting a negated finding can result in a harmful diagnosis. For example, misinterpreting ‘pleural effusion’ versus ‘no pleural effusion’ can lead to incorrect diagnoses with serious clinical consequences. 

The Stanford AIMI CheXagent model, while impressive in its ability to evaluate prompts in many different disciplines, modality, and diagnoses, falls short in recognizing hard negation due to semantic similarity. The high degree of overlap in wording and structure between positive and negative statements can cause the model to encode both into closely aligned text embeddings and misinterpret negated findings as positive ones. To confirm this observation, the model was rigorously tested for a specific diagnosis, pleural effusion. A dataset of 99 prompts was used that contained a diverse mix of positive and negative phrasing for the absence or presence of pleural effusion. We then fine-tuned the model, first using InfoNCE \cite{Authors14o} contrastive loss and then using the CoN-CLIP basis for fine-tuning. The three model checkpoints were then tested using token attribution and t-SNE tests to understand aspects of the model’s improvements.  This approach allowed us to compare how each fine-tuning strategy impacted the internal representation of the negation of the model and its ability to correctly retrieve clinically relevant images.
\section{Related Works}
\label{sec:formatting}

\subsection{Vision Language Models in Clinical Tasks}
Recent work "Bringing CLIP to the Clinic: Dynamic Soft Labels and Negation-Aware Learning for Medical Analysis" \cite{Authors14} has looked at how to adapt CLIP for medical imaging, especially for tasks like report matching and image retrieval. Ko and Park (2025) use 14 labels extracted from reports by CheXbert \cite{Authors14n} and InfoNCE \cite{Authors14o} for cross-modal alignment to propose a negation-aware training strategy by introducing hard negatives based on clinical contradictions such as “pleural effusion” versus “no pleural effusion.” Their model uses dynamic soft labels and graph-based alignment loss to improve retrieval accuracy, and explored measures beyond cosine similarity \cite{Authors14p}. Their results show that models trained with these negated contrasts can better distinguish between similar but semantically opposite findings.

\subsection{Location of Negation}
The paper "How and where does CLIP process negation?" \cite{Authors14b} by Quantmeyer et al. (2024) asks where and how CLIP processes negation internally. Their method adapted from Meng et al. \cite{Authors14q} use causal tracing to locate the specific transformer layers and attention heads that handle negation. They build on the VALSE benchmark \cite{Authors14r} and find that negation tends to be encoded most strongly in the middle layers of the text encoder. Validity is tested with a subset of the CANNOT dataset \cite{Authors14s}. Their work track how negation flows through a vision language model and shows that CLIP’s understanding of negation is not evenly distributed between various layers.

\subsection{CLIP in Clinical Tasks}
Chen et al. (2024) introduce CheXagent in "A Vision-Language Foundation Model to Enhance Efficiency of Chest X-ray Interpretation" \cite{Authors14c}, a CLIP-based foundation model trained on chest X-rays for classification, segmentation, and report generation. The language model was trained on clinical text \cite{Authors14u} and the image encoder was trained with SigLip \cite{Authors14t}. While it performs well on a wide range of tasks and addresses challenges shown in prior works \cite{Authors14v}, it doesn’t directly address negation.

\subsection{Use of distractors for contrastive learning}
A recent study by Singh et al. titled “Learn ‘No’ to Say ‘Yes’ Better” \cite{Authors14d} directly addresses a core limitation in CLIP and similar vision-language models (VLMs): the failure to understand and act on textual negation \cite{Authors14w}. They confirm that models like CLIP frequently misinterpret prompts with negations \cite{Authors14x}, especially with non fluid language. To resolve this, they propose CoN-CLIP, a fine-tuning framework that incorporates negated captions and visually anchored distractor images into the contrastive learning objective. They froze the
image encoder and fine-tune CLIP’s text encoder on the final loss function Lconclip, similar to \cite{Authors14y}. They achieved 99.7 \% accuracy on held-out CC-Neg samples. This work highlights the importance of training VLMs with high-quality negation examples to enhance semantic precision.

\section{Methods}

\subsection{Dataset Preparation and Ethics}

We obtained a dataset of 3,996 chest X-ray images and corresponding radiology reports from the Open-i repository \cite{Authors14e}. All data in Open-i are fully de-identified prior to public release and originate from open-access clinical collections, making it possible for non-commercial academic use without Institutional Review Board oversight. Images were first sorted into frontal and lateral views, and only frontal images were retained for consistency across cases. We then filtered for pleural effusion due to its clinical importance and frequent appearance in the dataset, producing a balanced training set of 230 images and a balanced test set of 70 images (50\% effusion, 50\% no effusion). All data were used in accordance with their respective open-access or Creative Commons licenses, and no patient-identifiable information was accessed.

\begin{figure*}[t]
  \centering
  \includegraphics[width=1\textwidth]{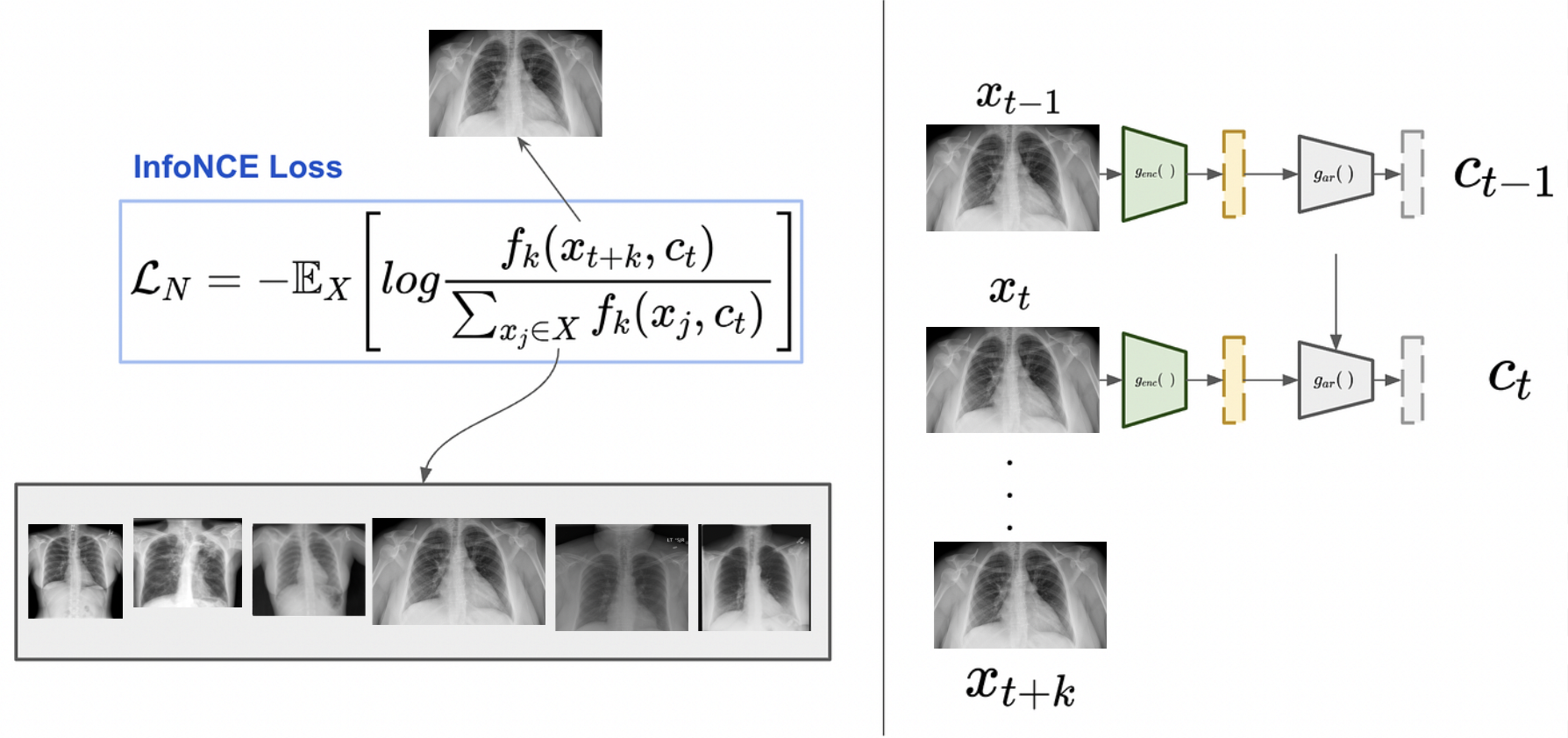}
  \caption{Overview of our contrastive setup, adapted from a diagram in a Medium article on InfoNCE loss\cite{medium_infonce}. All chest X-ray images shown in figures were obtained from the Open-i repository and originate from the PubMed Central Open Access subset or the Indiana University Chest X-ray dataset\cite{Authors14e}.}
  \label{fig:reunotes}
\end{figure*}

\subsection{Fine-tuning CON1 CLIP}

\begin{figure*}[t]
  \centering
  \includegraphics[width=1\textwidth]{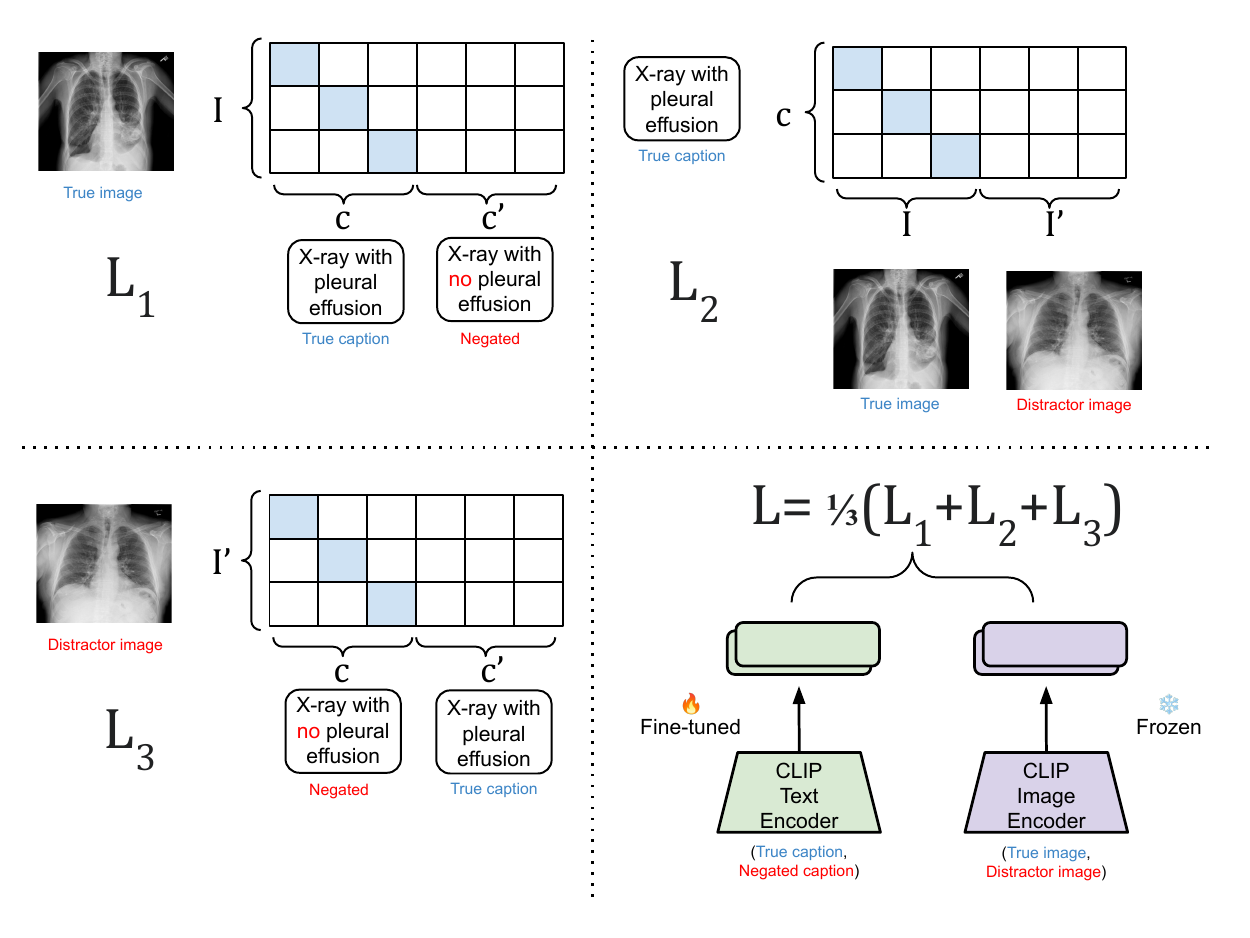}
  \caption{Use of negations and distractor images in a contrastive objective for finetuning CON2 CLIP text encoder for negation understanding adapted from Singh et al. \cite{Authors14d} All chest X-ray images shown in figures were obtained from the Open-i repository and originate from the PubMed Central Open Access subset or the Indiana University Chest X-ray dataset\cite{Authors14e}.}
  \label{fig:reunotes}
\end{figure*}

We fine-tuned the Stanford AIMI CheXagent model (ViT-B/16, LAION2B-s34B-b88K checkpoint) \cite{Authors14c} to better align its text encoder with clinically meaningful natural-language prompts (Figure 2). Despite its large radiology training set, CheXagent isn’t specifically trained to handle subtle differences caused by negation. Because our task required sensitivity to small changes in semantics, we froze the vision encoder to preserve the pretrained radiographic representations and fine-tuned only the text encoder. This ensured that the model’s image features remained stable while allowing the langauge model to adapt to the structure of our prompts.

Training followed a standard CLIP-style contrastive learning framework using the InfoNCE loss. For each batch of size $N$, we computed an $N \times N$ similarity matrix by taking the dot product between all L2-normalized image and text embeddings. The diagonal entries correspond to correct image--prompt pairs, while the off-diagonal entries served as negative examples. Since our dataset contained only positive matches, these in-batch negatives were necessary with the goal of mirroring the negative sampling strategy of the original CLIP training procedure\cite{Authors14f}.

A cross-entropy loss was applied across matrix rows, encouraging the model to assign the highest similarity score to each image's correct prompt. This is to sharpen the semantic separation between clinically distinct descriptions. Fine-tuning was performed for five epochs with a batch size of 32 using the AdamW optimizer and a learning rate of $1 \times 10^{-5}$, applied only to the text encoder. All embeddings were L2-normalized prior to similarity computation, and training was performed on a GPU when available. The resulting model is referred to as \textbf{CON1 CLIP}.

\subsection{Fine-tuning CON2 CLIP}

To address the model’s difficulty with clinical negation, we performed a second fine-tuning stage inspired by the CoN-CLIP framework introduced by (Singh et al.)\cite{Authors14d}. CoN-CLIP is designed to incorporate semantic opposition directly into the contrastive objective, making it usable for medical settings in which negation is frequently used for medical reports. We adapted the CoN-CLIP objective to the clinical domain, modifying the construction of negated prompts, distractor images, and batch structure to suit chest X-ray data rather than the caption–image pairs used in the original work. 

For each training example, we included a true chest X-ray image, a natural-language prompt describing its clinical finding, a distractor image matched to the opposite condition, and a distractor prompt expressing that opposite state. These components were generated from a curated CSV file that ensured semantic opposition between the true and distractor elements. This design allowed the model to repeatedly contrast ``effusion’’ against ``no effusion’’ during optimization, directly targeting the model’s known limitation in handling negation.

As with CON1 CLIP, the image encoder was frozen to preserve CheXagent’s pretrained visual features, and only the text encoder was fine-tuned (Figure 3). For each batch, we computed three contrastive losses following the CoN-CLIP formulation:
\begin{itemize}
    \item $L_1$: Align each true image with its corresponding prompt while repelling the distractor prompt.
    \item $L_2$: Align each true prompt with its corresponding image while repelling the distractor image.
    \item $L_3$: Align each distractor image with its own prompt while penalizing similarity to the true prompt.
\end{itemize}
All embeddings were L2-normalized, and cosine similarities were scaled by a temperature parameter $\tau = 0.07$. The final loss function was the mean of the three components:
\[
\mathcal{L}_{\text{conclip}} = \frac{L_1 + L_2 + L_3}{3}.
\]
Fine-tuning was performed for five epochs using a batch size of 32 and the AdamW optimizer with a learning rate of $1 \times 10^{-6}$. We used a smaller learning rate than in CON1 to stabilize training under the more complex multi-term objective. The resulting model, which incorporates explicit semantic opposition during training, is referred to as \textbf{CON2 CLIP}.

\section {Experimentation}
\subsection{Attribution}

Attribution analysis was used to investigate how each model handled negation tokens in the input prompts. We applied gradient-based attribution methods to compute the contribution of individual tokens to the image–text similarity score. We focused on the token containing the negation and measured its attribution across the base CheXagent model, our first fine-tuned model (CON1 CLIP), and the second fine-tuned model (CON2 CLIP) using the CoN-CLIP objective (Figure 4).
\begin{figure}[H]
    \centering
    \includegraphics[width=\linewidth]{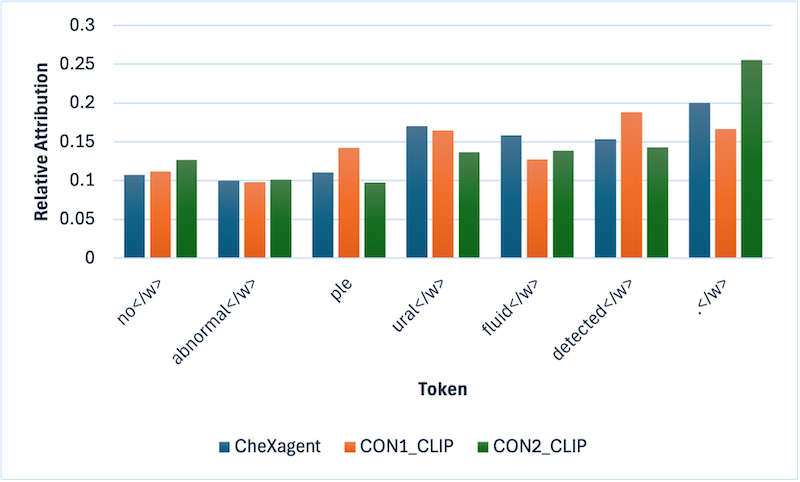}
    \caption{Token attribution of CheXagent, CON1 CLIP, and CON2 CLIP given a prompt containing negation for pleural effusion}
    \label{fig:tokenattribution}
\end{figure}

In the CheXagent model, the relative attribution to the negation token was low at 0.107, suggesting poor sensitivity to negation semantics. After fine-tuning with in-batch contrastive learning in CON1 CLIP, the attribution increased to 0.112, suggesting improved awareness of negation structure. The CON2 CLIP model showed the highest attribution (0.127), reflecting further gains achieved through training with opposed image-text pairs. These results show that training the model with opposite meanings helps it understand clinical negation better.

\subsection{t-SNE Analysis of Text Embedding Structure}
\begin{figure}[t]
    \centering
    \includegraphics[width=1\linewidth]{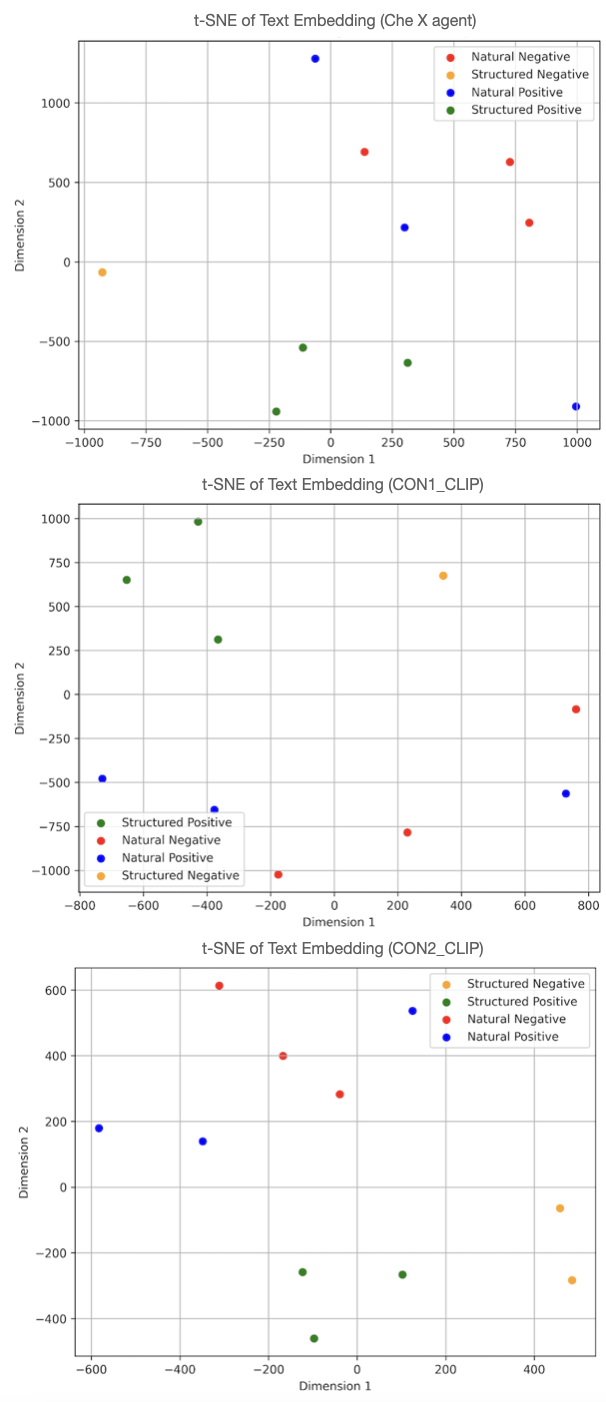}
    \caption{t-SNE graph of CheXagent (top), CON1 CLIP (middle), and CON2 CLIP (bottom) depicting Structured Positive, Structured Negative, Natural Positive, and Natural Negative prompt clusters.}
    \label{fig:chextsne}
\end{figure}
To investigate how each model organizes semantic information related to pleural effusion, we performed a t-SNE projection of text embeddings for natural positives, natural negatives, structured positives, and structured negatives (Figure 5). This analysis allowed us to visualize the geometry of the text embedding space and to assess how effectively each model separates negated and non-negated clinical statements.

In the base CheXagent model, the four categories showed loose and overlapping structure. Natural language prompts in particular exhibited substantial dispersion, with natural negatives often embedded near natural positives. This indicates that the pretrained model does not strongly encode the difference between “pleural effusion” and “no pleural effusion,” which is consistent with its weak attribution to the negation token and its poorer retrieval performance on negated prompts.

Fine-tuning with in-batch contrastive learning in CON1 CLIP improved cluster definition, especially for structured prompts. Structured positives and structured negatives became more clearly separated, suggesting that the model learned to use the quadriplet-derived format to distinguish between affirming and negating descriptions. However, natural language prompts remained relatively diffuse, and natural negatives continued to overlap with natural positives. This mismatch between structured and natural formats further supports that although CON1 CLIP increases the model’s sensitivity to negation, it still struggles to extend that understanding to more natural phrasing.

The CON2 CLIP model, trained with the CoN-CLIP objective and explicit semantic opposition, produced the most distinct and compact clusters across all four prompt types. Natural negatives became markedly more cohesive and separated from natural positives. Structured and natural prompts of the same polarity also lined up better, showing that CON2 CLIP learned a more consistent, format-independent understanding of negation.

The t-SNE plots reveal a shift from disorganized embeddings in the pretrained model to well-defined clusters in CON1 and especially CON2. This supports our retrieval and attribution findings, showing that contrastive training with semantically opposed examples improves the model’s grasp of clinical negation.

\subsection {Ablation Study}
To better understand how negation-related information is handled within the CLIP text encoder, we conducted an attention-head ablation analysis across the base CheXagent model, the contrastively fine-tuned CON1 CLIP model, and the CoN-CLIP fine-tuned CON2 CLIP model (Figure 6). This analysis isolates the contribution of individual attention heads to the image–text similarity score produced during retrieval for negated prompts.

For each model, we selected a representative negated prompt (“No evidence of pleural effusion.”) and a corresponding chest X-ray image from the test set. We first computed the baseline similarity score between the image and negated prompt. Then, for every layer–head pair in the 12-layer, 8-head ViT-B text transformer, we performed a targeted zero-ablation in which the output of that specific attention head was set to zero during the forward pass and the image–text similarity was recomputed. The change in similarity (sim) quantifies how much that head contributes to the model’s processing of negation in that prompt. Positive sim values indicate that removing the head reduces negation-related similarity, while negative values indicate the head interfered with negation alignment.

This produced a 12×8 matrix of sim scores for each model, which we visualized using heatmaps. These ablation maps allow us to characterize how dispersed or localized negation processing is across the transformer layers, and how fine-tuning redistributes responsibility for semantic opposition across the text encoder.

\begin{figure}[t]
    \centering
    \includegraphics[width=1\linewidth]{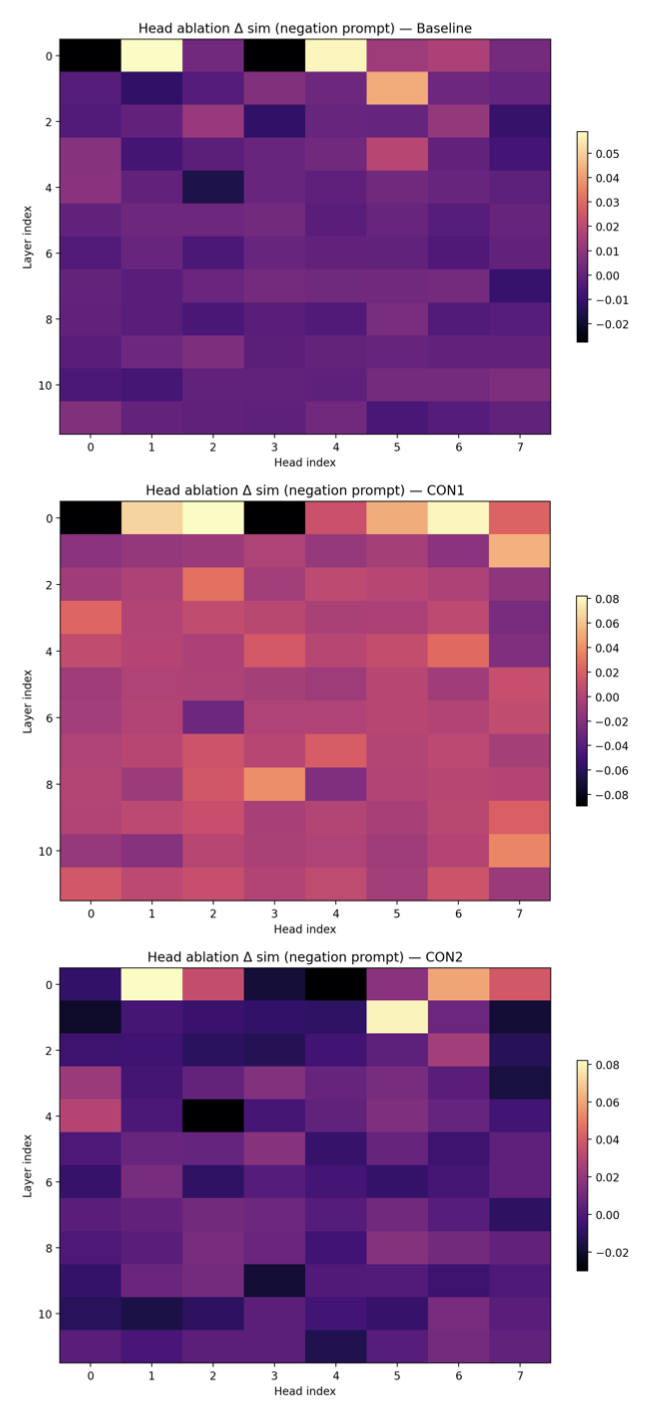}
    \caption{Attention-head ablation maps for CheXagent (top), CON1 CLIP (middle), and CON2 CLIP (bottom) showing how individual heads affect image–text similarity for a negated prompt.}
    \label{fig:chextsne}
\end{figure}

\section{Results and Analysis}

\begin{figure*}[t]
  \centering
  \includegraphics[width=1\textwidth]{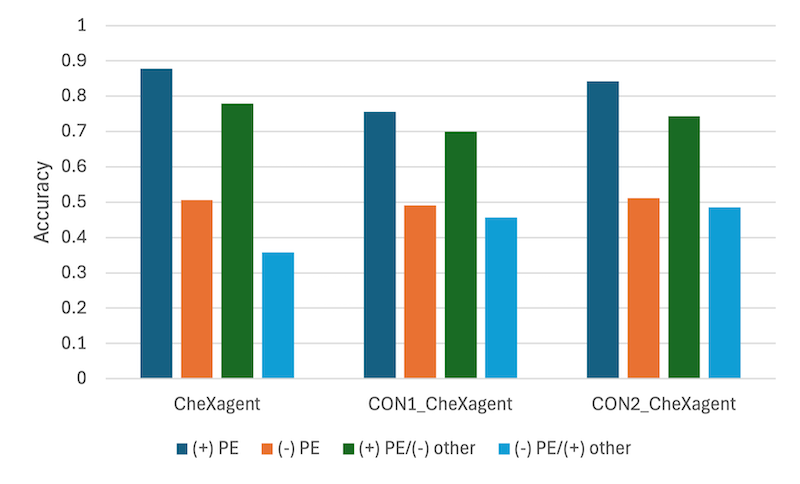}
  \caption{Accuracy results comparing CheXagent, CON1 CheXagent, and CON2 CheXagent from retrieving 10 images given four types of prompts across 99 prompts.}
  \label{fig:reunotes}
\end{figure*}

We evaluated all three models across attribution, embedding-space structure, retrieval accuracy, and attention-head ablations. Together, these analyses provide a multi-level view of how each fine-tuning strategy reshaped the model’s internal handling of clinical negation and how these changes translate into behavior relevant to real radiology.

\subsection{Retrieval Accuracy}

Top-10 retrieval accuracy was measured on a 70-image balanced test set using natural positives, natural negatives, structured positives, and structured negatives (Figure 7). The base CheXagent model showed the weakest performance on negated prompts, frequently retrieving images depicting pleural effusion even when the query explicitly negated it. This pattern mirrors clinical failure modes in which automated tools mistakenly interpret negated findings as positive.

Fine-tuning with standard in-batch contrastive learning (CON1 CLIP) improved retrieval accuracy for negated and composite prompts by approximately 10 \%, demonstrating increased sensitivity to the lexical cues associated with negation. However, CON1 exhibited a slight drop in accuracy for strictly positive prompts, suggesting that sharpening semantic contrast around negation sometimes narrows the model’s embedding-space tolerance for affirmative findings.

The CoN-CLIP-inspired model (CON2 CLIP) achieved the strongest performance. Retrieval accuracy for negated and composite prompts increased by nearly 15 \% relative to CheXagent, with a smaller decline on positive prompts compared to CON1. These results confirm that explicitly pairing each training sample with an opposed prompt and image leads to more reliable negation comprehension in a clinically meaningful task.

\subsection{Token-Level Attribution}

Gradient-based token attribution provided insight into how the models weight individual linguistic components during image–text matching. CheXagent assigned low attribution to the negation token (0.107), indicating that “no” contributed little to its final similarity computation. CON1 CLIP modestly increased attribution to 0.112, reflecting improved attention to negation structure. CON2 CLIP reached the highest attribution (0.127), demonstrating that explicit semantic opposition in training encourages the model to treat negation as a meaningful and influential modifier. This progression mirrors the improvements observed in retrieval accuracy, suggesting consistent gains across both token-level and model-level behavior.

\subsection{t-SNE Analysis of Text Embedding Structure}

t-SNE projections revealed how the text encoder organizes semantic information across structured and natural prompts. In CheXagent, clusters were poorly separated, in which natural negatives were often embedded near positive statements, showing that negation signals were not distinctly represented. CON1 CLIP improved separation for structured prompts but struggled to generalize this improvement to natural language prompts, suggesting overfitting to the quadriplet structure.

CON2 CLIP produced the most coherent embedding geometry with structured and natural negatives forming compact, distinct clusters that were well separated from their positive counterparts. This alignment across prompt formats show that semantic opposition was internalized more uniformly, allowing format consistent modeling of negation.

\subsection{Attention-Head Ablation Analysis}

Attention-head ablation further clarified how negation cues propagate through the encoder. The baseline CheXagent model show sparse and highly localized head contributions, which correlates with its weak negation attribution and disorganized t-SNE structure.

CON1 CLIP broadened the distribution of influential heads, demonstrating a more distributed processing pattern. Several heads also yielded negative $\Delta \mathrm{sim}$ scores, implying that parts of the network actively interfered with negation alignment.

CON2 CLIP exhibited the most organized ablation map. Heads showing positive $\Delta \mathrm{sim}$ values clustered in mid-to-late layers, suggesting deeper and more consistent processing of negation. These findings align with CON2’s improved attribution, t-SNE separation, and retrieval metrics.

Overall, the many analyses consistently demonstrate that the CoN-CLIP-inspired training produces the most robust and clinically reliable handling of negation, addressing a key limitation in current CLIP-based radiology models.

\section{Discussion}
This work demonstrates that standard CLIP-based models, including the widely used Stanford AIMI CheXagent, exhibit systematic failures in interpreting clinical negation. These failures pose direct risks in retrieval and clinical dependent applications, where confusing “pleural effusion” with “no pleural effusion” can lead to meaningful errors. These observations reinforce that negation must be handled reliably in any model intended for medical retrieval or reporting workflows.

Our findings show that fine-tuning the text encoder alone can meaningfully shift the model’s internal semantic structure without changing the vision encoder. CON1 CLIP, which used a standard in-batch contrastive objective, improved negation sensitivity and sharpened separation between structured prompts. It also revealed, however, inconsistencies across natural-language phrasing and introduced a small drop in accuracy for clearly positive prompts. This suggests that improving one clinically important capability may slightly degrade performance on others, and these trade-offs must be explicitly evaluated before clinical adoption.

The second approach, CON2 CLIP, directly incorporated semantic opposition through a CoN-CLIP-style formulation. This fine-tuning strategy yielded more stable negation-token attribution and more format-invariant t-SNE clusters. Retrieval accuracy for negated and composite prompts improved by nearly 15\% over the base model, which suggests a clinically relevant reduction in negation-related retrieval errors. CON2 CLIP also showed fewer unpredictable or “noisy’’ internal behaviors, which is a key requirement for clinical trust.

While fine-tuning for improved recognition of negation was successful, one thing to consider is that our datase is relatively small compared to the scale at which foundation models are usually adapted, and the distribution is limited to a single condition drawn from a single institution (OpenI). This raises concerns about generalization. 

Future work should extend these methods to additional thoracic findings and evaluate generalization across multiple datasets. Quantifying uncertainty and clinician-guided testing will be necessary in evaluating whether improved negation comprehension results in practical benefits in clinical settings. More broadly, increasing the semantic robustness of vision–language models is an important step toward creating AI systems that are safer and more trustworthy in real-world medical practice.

\section{Conclusion}
This study shows that standard CLIP-based models struggle to reliably interpret clinical negation. We introduced two targeted fine-tuning strategies that adapt the text encoder for improved handling of negated descriptions. CON1 CLIP achieved modest improvements using an in-batch contrastive objective, while CON2 CLIP, adapted from the CoN-CLIP framework, produced larger gains, clearer internal representations, and nearly a 15\% improvement in retrieval accuracy for negated prompts.

Although these improvements come with minor performance trade-offs for positive-only prompts, the results demonstrate that contrastive fine-tuning can substantially enhance the reliability of vision–language models for clinical tasks involving negation. Strengthening this capability brings retrieval systems closer to the consistency and safety expectations required for medical AI tools for future clinical use.
{
    \small
    \bibliographystyle{ieeenat_fullname}
    \bibliography{main}
}

\end{document}